\pdfoutput=1
\documentclass[11pt]{article}

\usepackage[]{ACL2023}

\usepackage{times}
\usepackage{latexsym}

\usepackage[T1]{fontenc}
\usepackage[utf8]{inputenc}

\usepackage{microtype}

\usepackage{inconsolata}

\usepackage[ruled, vlined, linesnumbered]{algorithm2e}
\usepackage{bm}
\usepackage{graphicx}
\usepackage{amsmath}
\usepackage{multirow}
\usepackage{cleveref}
\usepackage{xspace}
\usepackage{xcolor}
\usepackage{enumitem}
\usepackage{placeins}
\usepackage{soul}
\usepackage{bbm}
\usepackage{booktabs}
\usepackage{hyperref}
\usepackage{subcaption}
\usepackage{pgfplots}
\DeclareUnicodeCharacter{2212}{−}
\usepgfplotslibrary{groupplots,dateplot}
\usetikzlibrary{patterns,shapes.arrows}
\pgfplotsset{compat=newest}

\newcommand{\ours}{{MetaAdapt}\xspace}

\title{\ours: Domain Adaptive Few-Shot Misinformation Detection via Meta Learning}

\author{{Zhenrui Yue \quad Huimin Zeng \quad Yang Zhang \quad Lanyu Shang \quad Dong Wang} \\
  University of Illinois Urbana-Champaign \\
  \texttt{\{zhenrui3, huiminz3, yzhangnd, lshang3, dwang24\}@illinois.edu}}

\begin{document}
\maketitle
\begin{abstract}
With emerging topics (e.g., COVID-19) on social media as a source for the spreading misinformation, overcoming the distributional shifts between the original training domain (i.e., source domain) and such target domains remains a non-trivial task for misinformation detection. This presents an elusive challenge for early-stage misinformation detection, where a good amount of data and annotations from the target domain is not available for training. To address the data scarcity issue, we propose \ours, a meta learning based approach for domain adaptive few-shot misinformation detection. \ours leverages limited target examples to provide feedback and guide the knowledge transfer from the source to the target domain (i.e., learn to adapt). In particular, we train the initial model with multiple source tasks and compute their similarity scores to the meta task. Based on the similarity scores, we rescale the meta gradients to adaptively learn from the source tasks. As such, \ours can learn how to adapt the misinformation detection model and exploit the source data for improved performance in the target domain. To demonstrate the efficiency and effectiveness of our method, we perform extensive experiments to compare \ours with state-of-the-art baselines and large language models (LLMs) such as LLaMA, where \ours achieves better performance in domain adaptive few-shot misinformation detection with substantially reduced parameters on real-world datasets.
\end{abstract}

\section{Introduction}
\label{sec:intro}

\begin{figure}[t]
    \centering
    \includegraphics[trim=6.9cm 4.3cm 7.7cm 3.6cm, clip, width=0.9\linewidth]{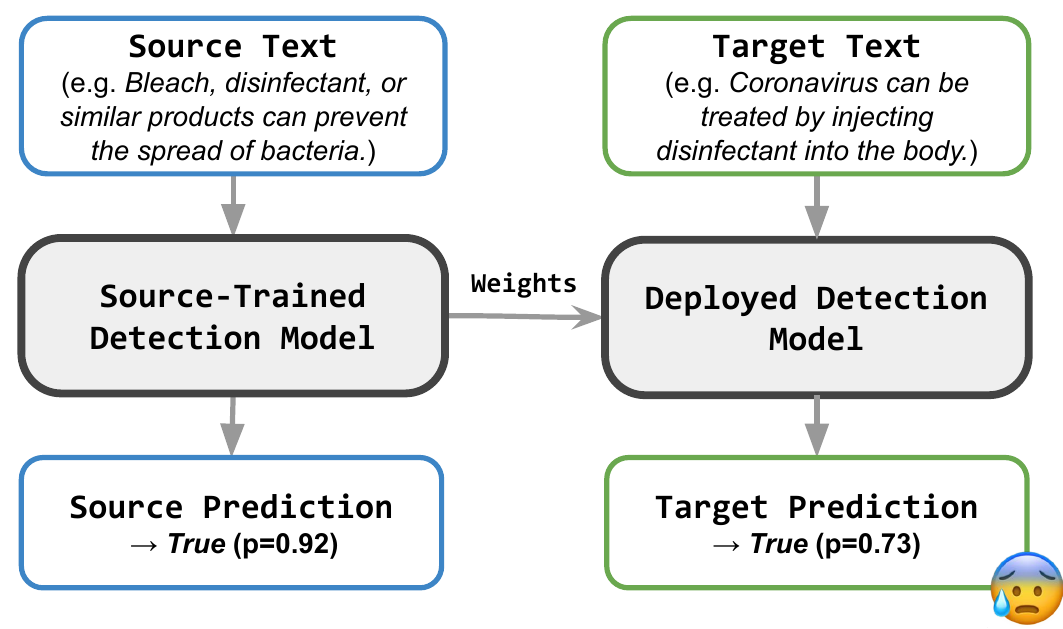}
    \caption{Existing models (from source domain) fail to detect rumors on emerging topics (target domain).} 
    \label{fig:intro}
    \vspace{-10pt}
\end{figure}

% misinformation detection
Recently, significant progress has been made in misinformation detection due to the improvements in developing machine learning-based methods~\cite{wu2019misinformation, shu2020disinformation, wu-etal-2022-cross}. Such methods include large language models (LLMs), which can be fine-tuned for detecting and responding to rumors on social media platforms~\cite{jiang2022fake, he2023reinforcement, touvron2023llama}. However, misinformation on emerging topics remains an elusive challenge for existing approaches, as there exists a large domain gap between the training (i.e., source domain) and the target distribution (i.e., target domain)~\cite{yue2022contrastive}. For instance, existing models often fail to detect early-stage misinformation due to the lack of domain knowledge (see \Cref{fig:intro}).

% problem and existing solutions
With the increase of emerging topics (e.g., COVID-19) on social media as a source of misinformation, the failure to distinguish such early-stage misinformation can result in potential threats to public interest~\cite{roozenbeek2020susceptibility, chen2022combating}. To tackle the problem of cross-domain early misinformation detection, one possible solution is crowdsourcing, which collects domain knowledge from online resources~\cite{medina-serrano-etal-2020-nlp, hu-etal-2021-compare, shang2022privacy, kou2022crowd}. Another alternative approach is to transfer knowledge from labeled source data to unlabeled target data with domain adaptive methods~\cite{zhang2020bdann, li2021multi, suprem2022midas, shu2022cross, yue2022contrastive, zeng2022unsupervised}. However, the former methods use large amounts of human annotations while the latter approaches require extensive unlabeled examples. As such, existing methods are less effective for detecting cross-domain early misinformation, where neither large amounts of annotations nor target domain examples can be provided for training.

% intro to few shot and meta learning
Despite the insufficiency of early misinformation data, limited target examples and annotations can often be achieved at minimal costs~\cite{kou2021fakesens, kou2022hc, shang2022knowledge}. Nevertheless, previous approaches are not optimized to learn from the source data under the guidance of limited target examples~\cite{zhang2020bdann, lee-etal-2021-towards, mosallanezhad2022domain, yue2022contrastive}. Such methods are often unaware of the adaptation objective and thus fail to maximize the transfer of source domain knowledge. To fully exploit existing data from different domains, we consider a cross-domain few-shot setting to adapt misinformation detection models to an unseen target domain~\cite{motiian2017few, zhao2021domain, lin-etal-2022-detect}. That is, given a source data distribution and access to limited target examples, our objective is to maximize the model performance in the target domain. An example of such application can be adapting a model from fake news detection to COVID early misinformation detection, where abundant fake news from existing datasets can be used for training the model under the guidance of limited COVID misinformation examples.

% our method
In this paper, we design \ours, a few-shot domain adaptation approach based on meta learning for early misinformation detection. Specifically, we leverage the source domain examples (i.e., source task) and train a model to obtain the task gradients. Then, we evaluate the updated model on the few-shot target examples (i.e., meta task) to derive second-order meta gradients w.r.t. the original parameters. We additionally compute the similarity between the task gradients and meta gradients to select more `informative' source tasks, such that the updated model adaptively learns (from the source data) to generalize even with a small number of labeled examples. In other words, the meta model learns to reweight the source tasks with the objective of optimizing the model performance in the target domain. Therefore, the resulting model can optimally adapt to the target distribution with the provided source domain knowledge. To show the efficacy of our meta learning-based adaptation method, we focus on the early-stage misinformation of COVID-19 and demonstrate the performance of \ours on real-world datasets, where \ours can consistently outperform state-of-the-art methods and large language models by demonstrating significant improvements.

We summarize our contributions as follows\footnote{We adopt publicly available datasets in the experiments and release our implementation at https://github.com/Yueeeeeeee/MetaAdapt.}:
\begin{enumerate}
\item We propose a few-shot setting for domain adaptive misinformation detection. Here, the labeled source data and limited target examples are provided for the adaptation process.
\item We propose \ours, a meta learning-based method for few-shot domain adaptive misinformation detection. Our \ours `learns to adapt' to the target data distribution with limited labeled examples.
\item \ours can adaptively learn from the source tasks by rescaling the meta gradients. Specifically, we compute similarity scores between the source and meta tasks to optimize the learning from the source distribution.
\item We show the effectiveness of \ours in domain adaptive misinformation detection on multiple real-world datasets. In our experiments, \ours consistently outperforms state-of-the-art baselines and LLMs.
\end{enumerate}
\section{Related Work}
\label{sec:related}

\subsection{Misinformation Detection}
% misinformation detection
Existing misinformation detection methods can be categorized into the following: (1)~content-based misinformation detection: such models are trained to perform misinformation classification upon input claims. For example, pretrained transformer models are used to extract semantic or syntactic properties to detect misinformation~\cite{karimi-tang-2019-learning, das2021heuristic, yue2022contrastive, jiang2022fake}. Moreover, multimodal input is used to learn text and image features that improve detection performance~\cite{khattar2019mvae, shang2021multimodal, santhosh2022multi, shang2022duo};
(2)~social-aware misinformation detection: user interactions can be used to evaluate online post credibility~\cite{jin2016news}. Similarly, patterns on propagation paths help detect misinformation on social media platforms~\cite{monti2019fake, shu2020hierarchical}. Social attributes like user dynamics enhance misinformation detection by introducing context~\cite{shu2019beyond}. Combined with content-based module, misinformation detection systems demonstrate improved accuracy~\cite{mosallanezhad2022domain, lin-etal-2022-detect}; 
(3)~knowledge-based misinformation detection: external knowledge can be leveraged as supporting features and evidence in fact verification and misinformation detection~\cite{vo-lee-2020-facts, liu-etal-2020-fine, brand2021bart}. Knowledge graphs or crowdsourcing approaches can derive additional information for explainability in misinformation detection~\cite{cui2020deterrent, hu-etal-2021-compare, koloski2022knowledge, kou2022hc, shang2022privacy, wu2022adversarial}. Yet existing methods focus on improving in-domain performance or explainability, few-shot misinformation detection in a cross-domain setting is not well researched. Hence, we study domain adaptive few-shot misinformation detection using content-based language models in our work.

\subsection{Domain Adaptive Learning}
% domain adaptation for cv
Domain adaptive learning aims to improve model generalization on an unseen domain given a labeled source domain. Such methods are primarily studied in image and text classification problems~\cite{li2018domain, kang2019contrastive, sicilia2021domain}. In image classification, existing methods minimize the representation discrepancy between source and target domains to learn domain-invariant features and transfer source knowledge to the target domain~\cite{kang2019contrastive, na2021fixbi}. Similarly, domain-adversarial and energy-based methods adopt additional critique, with which domain-specific features are regularized~\cite{sicilia2021domain, xie2022active}. Class-aware contrastive learning is proposed for fine-grained alignment, which regularizes the inter-class and intra-class distances to achieve domain-invariant yet class-separating features~\cite{li2018domain, shen2022connect}.

% domain adaptation for text classification
In text classification, various approaches are proposed to improve the target domain performance in cross-domain settings~\cite{silva2021embracing, li2021multi, ryu2022knowledge, nan-etal-2022-improving}. For instance, domain-adversarial training is used to learn generalizable features to detect cross-domain multimodal misinformation~\cite{wang2018eann, lin-etal-2022-detect, shu2022cross}. Reinforcement learning and contrastive adaptation are also adopted for fine-grained domain adaptation in misinformation detection~\cite{mosallanezhad2022domain, yue2022contrastive}. Nevertheless, domain-adaptive misinformation detection is not well studied in the few-shot learning setting. Therefore, we combine both settings and develop a method tailored for few-shot domain adaptation in misinformation detection: \ours. By leveraging knowledge transfer via the proposed meta objective, our approach shows significant improvements on out-of-domain misinformation using only a few labeled examples.

\subsection{Few-Shot Learning}
% meta & few-shot learning
Few-shot learning aims to learn a new task with a few labeled examples~\cite{wang2020generalizing}. Existing few-shot learning approaches (e.g., prototypical networks) learn class-wise features in the metric space to rapidly adapt to new tasks~\cite{vinyals2016matching, snell2017prototypical}. Meta learning methods search for the optimal initial parameters for unseen few-shot tasks via second-order optimization~\cite{finn2017model, rajeswaran2019meta, zhou2021task}. In computer vision, few-shot domain adaptation is studied in image classification to transfer knowledge to an unseen target domain~\cite{motiian2017few, tseng2019cross, zhao2021domain}. For language problems, meta learning is proposed to improve the few-shot performance in language modeling and misinformation detection~\cite{sharaf-etal-2020-meta, han-etal-2021-meta, salem2021meta, zhang2021learning, lei-etal-2022-adaptive}. 

To the best of our knowledge, few-shot domain adaptive misinformation detection via meta learning is not studied in current literature. Moreover, the mentioned few-shot setting can be helpful in real-world scenarios (e.g., detecting rumors on emerging topics). As such, we propose meta learning-based \ours for misinformation detection. \ours leverages limited target examples and adaptively exploits the source domain knowledge via task similarity, and thus improves the few-shot misinformation detection performance in the unseen target domain.

\begin{figure*}[t]
    \centering
    \includegraphics[trim=3cm 3.5cm 3cm 3.5cm, clip, width=0.85\linewidth]{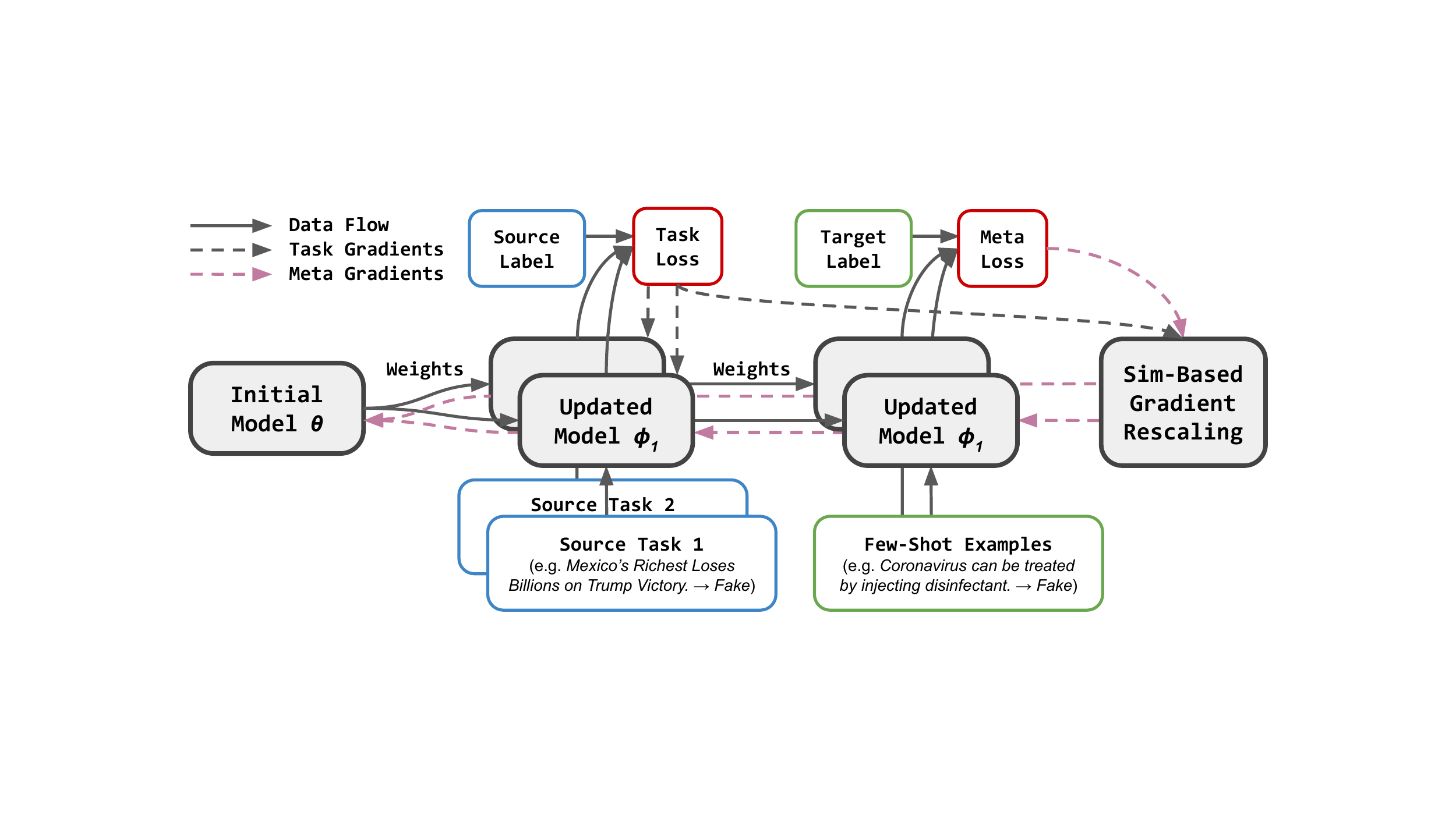}
    \caption{The proposed \ours, we illustrate the computation of task and meta gradients via task similarity.}
    \label{fig:method}
    \vspace{-5pt}
\end{figure*}

\section{Preliminary}
% problem
We consider the following problem setup for domain adaptive few-shot misinformation detection: labeled source data and $k$-shot target examples (i.e., $k$ examples per class) are available for training. The objective of our framework is to train a misinformation detection model $\bm{f}$ that is optimized for the target domain performance using both source and few-shot target examples.  

% data
\noindent
\textbf{Data}: Our research is defined within the scope of \emph{single-source} adaptive misinformation detection (i.e., we study the adaptation problem from a single source domain to the target domain). We denote $\bm{\mathcal{D}}_{s}$ as the source domain and $\bm{\mathcal{D}}_{t}$ as the (different) target domain. In our setting, labeled source data and limited target examples can be used for training. The few-shot adaptation is performed in two-fold: (1)~an initial model is updated upon multiple batches of sampled source data examples (i.e., source tasks) respectively; (2)~the updated models are evaluated on the few-shot target examples respectively (i.e., meta task) to compute the meta loss, followed by updating the initial parameters using the derived second-order derivatives. The input data is defined as follows:
\begin{itemize}[leftmargin=10pt]
    \item \emph{Labeled source data}: source training data $\bm{X}_{s}$ is provided by source domain $\bm{\mathcal{D}}_{s}$. Here, each sample $(\bm{x}_{s}^{(i)}, y_{s}^{(i)}) \in \bm{X}_{s}$ is a tuple comprising of input text $\bm{x}_{s}^{(i)}$ and label $y_{s}^{(i)} \in \{ 0, 1 \}$ (i.e., false or true). During training, source data batches are sampled as different `source tasks' and used to optimize the initial model.
    \item \emph{Few-shot target data}: we assume limited access to the target domain $\bm{\mathcal{D}}_{t}$. In other words, only $k$-shot subset $\bm{X}'_{t}$ from $\bm{X}_{t}$ is provided for training. Target samples are provided in the same label space with $(\bm{x}_{t}^{(i)}, y_{t}^{(i)}) \in \bm{X}'_{t}$, while the size of $\bm{X}'_{t}$ is constrained with $k$ examples in each label class (i.e., 10 in our experiments). $\bm{X}'_{t}$, or `meta task' is used to compute the meta loss and meta gradients w.r.t. the initial parameters.
\end{itemize}

% model
\noindent
\textbf{Model \& Objective:} The misinformation detection model is represented by a function $\bm{f}$ parameterized by $\bm{\theta}$. $\bm{f}$ takes textual statements as input and predicts the probability of input as true information, i.e., $y^{(i)} = \arg \max (\bm{f}(\bm{\theta}, \bm{x}^{(i)}))$. For optimization, our objective is to maximize the model performance on target data $\bm{X}_{t}$ (Note $\bm{X}_{t} \neq \bm{X}'_{t}$) from the target domain $\bm{\mathcal{D}}_{t}$. Mathematically, this can be formulated as the optimization problem of minimizing the loss $\mathcal{L}$ of $\bm{\theta}$ over target data $\bm{X}_{t}$:
\begin{equation}
    \min_{\substack{\bm{\theta}}} \mathcal{L}(\bm{\theta}, \bm{X}_{t}),
\end{equation}
where $\mathcal{L}$ is the loss function (i.e., cross-entropy).

\section{Methodology}
% some intro
Provided with labeled source data and $k$-shot target examples, we first present our meta adaptation framework for domain adaptive few-shot misinformation detection. To improve the adaptation performance, we introduce a second-order meta learning algorithm \ours based on learnable learning rate and task similarity. An illustration of \ours is provided in \Cref{fig:method}. Upon deployment, the adapted models achieve considerable improvements thanks to the adaptive optimization and similarity-guided meta adaptation.

\subsection{Few-Shot Meta Adaptation}
% formulation
Given model $\bm{f}$ with initial parameters $\bm{\theta}$, source dataset $\bm{X}_{s}$, few-shot target data $\bm{X}'_{t}$ and the number of tasks $n$ in each iteration, we formulate the meta adaptation framework as a bi-level optimization problem and provide a mathematical formulation:
\begin{equation}
    \label{eq:outer_lever}
    \min_{\substack{\bm{\theta}}} \frac{1}{n} \sum^{n} \mathcal{L}(\mathcal{A}lg(\bm{\theta}, \texttt{Sampler}(\bm{X}_{s})), \bm{X}'_{t}),
\end{equation}
in which \texttt{Sampler} stands for the source task sampler that draws source tasks of a fixed size from $\bm{X}_{s}$, $\mathcal{A}lg$ represents the optimization algorithm using first-order gradient descent, i.e.:
\begin{equation}
    \label{eq:inner_lever}
    \mathcal{A}lg(\bm{\theta}, \bm{X}) = \bm{\phi} = \bm{\theta} - \alpha \nabla_{\bm{\theta}} \mathcal{L}(\bm{\theta}, \bm{X}),
\end{equation}
with $\alpha$ representing the task learning rate and $\bm{\phi}$ representing the learnt parameter set over $\bm{X}$. 

% formulation explanation
In~\Cref{eq:outer_lever}, we are interested in learning an optimal parameter set $\bm{\theta}$ that minimizes the meta loss on the few-shot target set $\bm{X}'_{t}$, which can be denoted as the outer-level optimization of meta adaptation. The outer-level learning is achieved by deriving gradients w.r.t. $\bm{\theta}$ based on the meta loss using task-specific parameters (i.e., $\mathcal{A}lg(\bm{\theta}, \texttt{Sampler}(\bm{X}_{s}))$ or $\bm{\phi}$) and few-shot examples $\bm{X}'_{t}$. To obtain task-specific parameters, we sample a batch of source examples using \texttt{Sampler} and perform gradient descent steps on the original $\bm{\theta}$ (i.e, \Cref{eq:inner_lever}), which is known as the inner-level optimization. The inner-level optimization only requires first-order derivatives, however, to optimize the outer-level problem, it is necessary to differentiate through $\mathcal{A}lg$ (i.e., $\bm{\phi}$), which requires using second-order gradients~\cite{finn2017model}.

% some analysis
We now take a closer look at how to compute the derivatives with chain rule in meta adaptation:
\begin{equation}
  \begin{aligned}
    \label{eq:analysis}
    \frac{d \mathcal{L}(\mathcal{A}lg(\bm{\theta}, \bm{X}), \bm{X}'_{t})}{d \bm{\theta}} &= \\
    \frac{d\mathcal{A}lg(\bm{\theta}, \bm{X})}{d\bm{\theta}} \nabla_{\bm{\phi}} &\mathcal{L}(\mathcal{A}lg(\bm{\theta}, \bm{X}), \bm{X}'_{t}),
  \end{aligned}
\end{equation}
Note that $\mathcal{A}lg(\bm{\theta}, \bm{X})$ is equivalent to $\bm{\phi}$. The right-side component $\nabla_{\bm{\phi}}\mathcal{L}(\mathcal{A}lg(\bm{\theta}, \bm{X}), \bm{X}'_{t})$ refers to first-order derivatives by computing the meta loss using the few-shot examples $\bm{X}'_{t}$ and task-specific parameter set $\bm{\phi}$ ($\mathcal{L} \rightarrow \bm{\phi}$). This step can be computed with conventional gradient descent algorithms. The left-side component $\frac{d\mathcal{A}lg(\bm{\theta}, \bm{X})}{d\bm{\theta}}$ is a non-trivial step as it requires second-order derivatives (i.e., Hessian matrix) to track parameter-to-parameter changes through $\mathcal{A}lg(\bm{\theta}, \bm{X})$ to $\bm{\theta}$. In our implementation, we compute the meta gradients w.r.t. $\bm{\theta}$ using the meta evaluation loss similar to model agnostic meta learning~\cite{finn2017model}. We also adopt adaptive learning rate $\alpha$ and $\beta$ and cosine annealing to improve the convergence of source and meta tasks~\cite{antoniou2018train}.

\subsection{The Proposed \ours}
% intro
While meta adaptation leverage source tasks to improve the target domain performance, it learns homogeneously from all source tasks without considering the informativeness of each individual task. To further exploit the source domain knowledge, we propose a task similarity-based \ours for domain adaptive few-shot misinformation detection. We first estimate the task-specific parameters with adaptive learning rates, followed by rescaling the meta loss using the task similarity scores. The proposed method selectively learns from source tasks, and thus, further improves the adaptation performance. We present the training details of \ours in \Cref{alg:maml}.

% the computation of task and meta gradients
We initialize the model and denote the parameters with $\bm{\theta}$. For source task $i$, the original parameters are updated with first-order derivatives as we sample source tasks from $\bm{X}_{s}$. Specifically in each step, the task gradients can be computed with $\nabla_{\bm{\theta}} \mathcal{L}(\bm{\theta}, \texttt{Sampler}(\bm{X}_{s}))$, as in \Cref{eq:inner_lever}. After a few gradient descent steps, the parameters converge locally and we denote the updated parameters with $\bm{\phi}_{i}$. As we update multiple times in each source task, we denote the task gradients with $\bm{\phi}_{i} - \bm{\theta}$ for simplicity. Subsequently, the meta loss $\mathcal{L}(\bm{\phi}_{i}, \bm{X}'_{t})$ is computed using $\bm{\phi}_{i}$ and the few-shot target examples $\bm{X}'_{t}$. To compute the meta gradients with backpropagation, we follow the chain rule and compute the derivatives w.r.t. the original parameter set $\bm{\theta}$. Similar to \Cref{eq:analysis}, we use $\frac{d\bm{\phi}_{i}}{d\bm{\theta}} \nabla_{\bm{\phi}_{i}}\mathcal{L}(\bm{\phi}_{i}, \bm{X}'_{t})$ to denote the meta gradients.

\begin{algorithm}[t]
    \small
    \caption{\ours Algorithm}
    \label{alg:maml}
    \SetAlgoLined
    \textbf{Input}
    Parameter set $\bm{\theta}$, source data $\bm{X}_{s}$, $k$-shot data $\bm{X}'_{t}$, number of iterations $N$, number of tasks $n$\;
    \For{$\mathrm{iter} \in \{ 1, 2, ..., N \}$}{ 
        \For{$i \in \{ 1, ..., n \}$}{ 
            % \tcp{Inner-Level Optimization}
            Sample source task from $\bm{X}_{s}$\;
            Update task parameter set $\bm{\phi}_{i}$ with \Cref{eq:inner_lever}\;
            Compute meta loss and meta gradients using $\bm{\phi}_{i}$, $\bm{X}'_{t}$ as in \Cref{eq:analysis}\;
            Calculate similarity score $s_{i}$ with \Cref{eq:similarity}\;
        }
        % \tcp{Outer-Level Optimization}
        Normalize similarity scores $\bm{s}$ with \Cref{eq:rescale}\;
        Update original parameter $\bm{\theta}$ with \Cref{eq:update}\;
    }
\end{algorithm}

% task similarity scores
To compute task similarity scores, we leverage task and meta gradients. The objective of computing gradient similarity is to selectively learn from the source tasks. If the task and meta gradients yield a high similarity score, the parameters are converging to the same direction in both inner- and outer-loop optimization. Thus, the source task optimization path is more `helpful' to improve the meta task performance (i.e., target domain performance). Or if, on the contrary, then the source task may be less effective for improving the meta task performance. Based on this principle, we compute task similarity score $s_{i}$ with cosine similarity:
\begin{equation}
    \label{eq:similarity}
    s_{i} = \texttt{CosSim}(\bm{\phi}_{i} - \bm{\theta}, \frac{d\bm{\phi}_{i}}{d\bm{\theta}} \nabla_{\bm{\phi}_{i}}\mathcal{L}(\bm{\phi}_{i}, \bm{X}'_{t})).
\end{equation}
In each iteration, we sample $n$ source tasks and compute the similarity scores for each pair of task and meta gradients. Then, the similarity scores $[s_{1}, s_{2} , ..., s_{n}]$ are transformed to a probability distribution using tempered softmax:
\begin{equation}
    \label{eq:rescale}
    \bm{s} = \texttt{softmax}([\frac{s_{1}}{\tau}, \frac{s_{2}}{\tau}, ..., \frac{s_{n}}{\tau}]),
\end{equation}
where $\tau$ is the temperature hyperparameter to be selected empirically. Finally, we update the original parameters with rescaled meta gradients:
\begin{equation}
    \label{eq:update}
    \bm{\theta} - \beta \sum_{i}^{n} s_{i} \cdot \frac{d\bm{\phi}_{i}}{d\bm{\theta}} \nabla_{\bm{\phi}_{i}}\mathcal{L}(\bm{\phi}_{i}, \bm{X}'_{t}).
\end{equation}

% overall and novelty
In summary, \ours computes task and meta gradients using sampled source tasks and few-shot target examples. Then, task similarity is computed to find more `informative' source tasks, followed by tempered rescaling of the meta gradients. Finally, the updated model parameters should exploit the source domain knowledge and demonstrate improved performance on the target data distribution. The overall framework of \ours is illustrated in \Cref{fig:method}. Unlike previous works~\cite{motiian2017few, tseng2019cross, zhao2021domain, yue2022contrastive}, we discard domain-adversarial or feature regularization methods, instead, we propose to leverage meta adaptation to guide the knowledge transfer from the source to target domain. Additionally, similarity-based gradients rescaling is designed to exploit different source tasks to achieve fine-grained adaptation performance.
\section{Experiments}

\subsection{Settings}
\noindent
\textbf{Model}: Similar to~\cite{li2021multi, yue2022contrastive}, we select RoBERTa as the base model to encode input examples in \ours. RoBERTa is a transformer model pretrained on a variety of NLP tasks before the COVID pandemic~\cite{liu2019roberta}.

\noindent
\textbf{Evaluation}: To validate the proposed method, we follow~\cite{kou2022hc, li2021multi, yue2022contrastive} and split the datasets into training, validation and test sets. The few-shot target examples are selected as the first $k$ examples in the validation set and the rest validation examples are used for validating the model. For evaluation metrics, we adopt balance accuracy (BA), accuracy (Acc.) and F1 score (F1) to evaluate the performance. See evaluation details in \Cref{sec:implementation}.

\begin{table}[t]
\small
\centering
\begin{tabular}{@{}llccc@{}}
\toprule
\textbf{Domain}                  & \textbf{Dataset} & \textbf{\, BA $\uparrow$ \,} & \textbf{\, Acc. $\uparrow$ \,} & \textbf{\, F1 $\uparrow$ \,} \\ \midrule
\multirow{5}{*}{\textbf{Source}} & FEVER            & 0.796                 & 0.796                   & 0.817                 \\
                                 & GettingReal      & 0.846                 & 0.959                   & 0.978                 \\
                                 & GossipCop        & 0.776                 & 0.869                   & 0.917                 \\
                                 & LIAR             & 0.607                 & 0.632                   & 0.712                 \\
                                 & PHEME            & 0.863                 & 0.867                   & 0.898                 \\ \midrule
\multirow{3}{*}{\textbf{Target}} & CoAID            & 0.889                 & 0.972                   & 0.985                 \\
                                 & Constraint       & 0.970                 & 0.971                   & 0.973                 \\
                                 & ANTiVax          & 0.932                 & 0.921                   & 0.931                 \\ \bottomrule
\end{tabular}
\caption{Supervised experiment results. The upper and lower parts report source and target dataset performance.}
\label{tab:supervised}
\vspace{-10pt}
\end{table}

\begin{table*}[t]
\small
\centering
\setlength\tabcolsep{2pt}
\centerline{
\begin{tabular}{@{}llccccccccc@{}}
\toprule
\multirow{2}{*}{\textbf{Source}}      & \textbf{Target} & \multicolumn{3}{c}{\textbf{CoAID (2020)}}           & \multicolumn{3}{c}{\textbf{Constraint (2021)}}      & \multicolumn{3}{c}{\textbf{ANTiVax (2022)}}         \\ \cmidrule(l){2-11} 
                                      & \textbf{Metric} & \textbf{BA $\uparrow$} & \textbf{Acc. $\uparrow$} & \textbf{F1 $\uparrow$} & \textbf{BA $\uparrow$} & \textbf{Acc. $\uparrow$} & \textbf{F1 $\uparrow$} & \textbf{BA $\uparrow$} & \textbf{Acc. $\uparrow$} & \textbf{F1 $\uparrow$} \\
\midrule
\multirow{6}{*}{\textbf{FE}}          & Na{\"i}ve       & 0.636          & 0.928          & 0.962          & 0.501          & 0.524          & 0.687          & 0.559          & 0.627          & 0.741          \\
                                      & CANMD           & 0.626          & 0.918          & 0.956          & 0.684          & 0.683          & 0.686          & 0.650          & 0.679          & 0.749          \\
                                      & ACLR            & 0.721          & \ul{0.935}     & \ul{0.965}     & 0.648          & 0.651          & 0.697          & 0.739          & 0.758          & 0.805          \\
                                      & ProtoNet        & 0.751          & 0.869          & 0.925          & 0.784          & 0.788          & \ul{0.812}     & 0.748          & 0.716          & 0.718          \\
                                      & MAML            & \ul{0.780}     & \textbf{0.939} & \textbf{0.967} & \ul{0.812}     & \ul{0.808}     & 0.797          & \ul{0.826}     & \ul{0.808}     & \ul{0.823}     \\
                                      & Ours            & \textbf{0.829}$_{\small{\pm\text{0.020}}}$ & 0.875$_{\small{\pm\text{0.049}}}$ & 0.927$_{\small{\pm\text{0.031}}}$ 
                                                        & \textbf{0.828}$_{\small{\pm\text{0.001}}}$ & \textbf{0.826}$_{\small{\pm\text{0.001}}}$ & \textbf{0.829}$_{\small{\pm\text{0.004}}}$ 
                                                        & \textbf{0.868}$_{\small{\pm\text{0.025}}}$ & \textbf{0.880}$_{\small{\pm\text{0.036}}}$ & \textbf{0.904}$_{\small{\pm\text{0.037}}}$ \\
\midrule
\multirow{6}{*}{\textbf{GR}}          & Na{\"i}ve       & 0.574          & 0.920          & 0.958          & 0.500          & 0.503          & 0.670          & 0.558          & 0.627          & 0.741          \\
                                      & CANMD           & 0.669          & \ul{0.935}     & \ul{0.965}     & 0.744          & 0.742          & 0.737          & 0.582          & 0.632          & 0.729          \\
                                      & ACLR            & 0.693          & 0.928          & 0.961          & 0.683          & 0.689          & 0.736          & 0.660          & 0.695          & 0.766          \\
                                      & ProtoNet        & 0.720          & 0.639          & 0.757          & 0.672          & 0.664          & 0.608          & 0.736          & 0.756          & 0.804          \\
                                      & MAML            & \ul{0.813}     & \textbf{0.937} & \textbf{0.965} & \ul{0.808}     & \ul{0.803}     & \ul{0.786}     & \ul{0.819}     & \ul{0.802}     & \ul{0.819}     \\
                                      & Ours            & \textbf{0.830}$_{\small{\pm\text{0.062}}}$ & 0.928$_{\small{\pm\text{0.004}}}$ & 0.960$_{\small{\pm\text{0.003}}}$ 
                                                        & \textbf{0.819}$_{\small{\pm\text{0.012}}}$ & \textbf{0.819}$_{\small{\pm\text{0.010}}}$ & \textbf{0.823}$_{\small{\pm\text{0.006}}}$ 
                                                        & \textbf{0.886}$_{\small{\pm\text{0.035}}}$ & \textbf{0.882}$_{\small{\pm\text{0.042}}}$ & \textbf{0.902}$_{\small{\pm\text{0.043}}}$ \\
\midrule
\multirow{6}{*}{\textbf{GC}}          & Na{\"i}ve       & 0.612          & 0.927          & 0.961          & 0.513          & 0.536          & 0.693          & 0.561          & 0.629          & 0.742          \\
                                      & CANMD           & 0.685          & \ul{0.931}     & \ul{0.963}     & 0.802          & 0.803          & \ul{0.817}     & 0.761          & 0.777          & 0.823          \\
                                      & ACLR            & 0.687          & \textbf{0.933} & \textbf{0.964} & 0.712          & 0.715          & 0.744          & 0.811          & 0.809          & \ul{0.835}     \\
                                      & ProtoNet        & 0.708          & 0.609          & 0.731          & 0.786          & 0.782          & 0.770          & 0.730          & 0.715          & 0.736          \\
                                      & MAML            & \ul{0.816}     & 0.926          & 0.959          & \ul{0.813}     & \ul{0.809}     & 0.801          & \ul{0.826}     & \ul{0.810}     & 0.826          \\
                                      & Ours            & \textbf{0.824}$_{\small{\pm\text{0.026}}}$ & 0.918$_{\small{\pm\text{0.004}}}$ & 0.954$_{\small{\pm\text{0.002}}}$ 
                                                        & \textbf{0.826}$_{\small{\pm\text{0.023}}}$ & \textbf{0.826}$_{\small{\pm\text{0.023}}}$ & \textbf{0.833}$_{\small{\pm\text{0.023}}}$ 
                                                        & \textbf{0.896}$_{\small{\pm\text{0.001}}}$ & \textbf{0.907}$_{\small{\pm\text{0.000}}}$ & \textbf{0.930}$_{\small{\pm\text{0.000}}}$ \\
\midrule
\multirow{6}{*}{\textbf{LI}}          & Na{\"i}ve       & 0.640          & 0.926          & 0.960          & 0.516          & 0.538          & 0.693          & 0.558          & 0.626          & 0.741          \\
                                      & CANMD           & 0.770          & 0.894          & 0.940          & \ul{0.815}     & \ul{0.814}     & \ul{0.818}     & 0.755          & 0.784          & \ul{0.834}     \\
                                      & ACLR            & 0.766          & \ul{0.938}     & \ul{0.966}     & 0.756          & 0.760          & 0.786          & 0.805          & 0.793          & 0.814          \\
                                      & ProtoNet        & 0.793          & 0.910          & 0.950          & 0.738          & 0.746          & 0.788          & 0.599          & 0.576          & 0.581          \\
                                      & MAML            & \ul{0.813}     & \textbf{0.938} & \textbf{0.966} & 0.813          & 0.809          & 0.800          & \ul{0.824}     & \ul{0.807}     & 0.824          \\
                                      & Ours            & \textbf{0.815}$_{\small{\pm\text{0.031}}}$ & 0.910$_{\small{\pm\text{0.014}}}$ & 0.949$_{\small{\pm\text{0.008}}}$ 
                                                        & \textbf{0.820}$_{\small{\pm\text{0.008}}}$ & \textbf{0.820}$_{\small{\pm\text{0.006}}}$ & \textbf{0.828}$_{\small{\pm\text{0.002}}}$ 
                                                        & \textbf{0.873}$_{\small{\pm\text{0.026}}}$ & \textbf{0.883}$_{\small{\pm\text{0.036}}}$ & \textbf{0.906}$_{\small{\pm\text{0.038}}}$ \\
\midrule
\multirow{6}{*}{\textbf{PH}}          & Na{\"i}ve       & 0.622          & 0.929          & 0.962          & 0.502          & 0.526          & 0.688          & 0.558          & 0.627          & 0.742          \\
                                      & CANMD           & 0.531          & 0.938          & 0.967          & 0.559          & 0.565          & 0.624          & 0.653          & 0.676          & 0.740          \\
                                      & ACLR            & 0.709          & \ul{0.939}     & \ul{0.967}     & 0.716          & 0.719          & 0.746          & 0.733          & 0.754          & 0.804          \\
                                      & ProtoNet        & 0.721          & 0.780          & 0.867          & 0.693          & 0.686          & 0.644          & 0.628          & 0.635          & 0.687          \\
                                      & MAML            & \ul{0.780}     & \textbf{0.939} & \textbf{0.967} & \ul{0.816}     & \ul{0.812}     & \ul{0.802}     & \ul{0.819}     & \ul{0.805}     & \ul{0.823}     \\
                                      & Ours            & \textbf{0.828}$_{\small{\pm\text{0.028}}}$ & 0.909$_{\small{\pm\text{0.009}}}$ & 0.949$_{\small{\pm\text{0.005}}}$ 
                                                        & \textbf{0.818}$_{\small{\pm\text{0.012}}}$ & \textbf{0.818}$_{\small{\pm\text{0.012}}}$ & \textbf{0.828}$_{\small{\pm\text{0.013}}}$ 
                                                        & \textbf{0.896}$_{\small{\pm\text{0.016}}}$ & \textbf{0.880}$_{\small{\pm\text{0.032}}}$ & \textbf{0.902}$_{\small{\pm\text{0.030}}}$ \\
\midrule
\multirow{6}{*}{\textbf{Avg}}         & Na{\"i}ve       & 0.617          & 0.926          & 0.961          & 0.506          & 0.525          & 0.686          & 0.559          & 0.627          & 0.741          \\
                                      & CANMD           & 0.656          & 0.923          & 0.958          & 0.721          & 0.721          & 0.736          & 0.680          & 0.710          & 0.775          \\
                                      & ACLR            & 0.715          & \ul{0.935}     & \ul{0.965}     & 0.703          & 0.707          & 0.742          & 0.750          & 0.762          & 0.805          \\
                                      & ProtoNet        & 0.739          & 0.761          & 0.846          & 0.735          & 0.733          & 0.724          & 0.688          & 0.679          & 0.705          \\
                                      & MAML            & \ul{0.800}     & \textbf{0.936} & \textbf{0.965} & \ul{0.813}     & \ul{0.808}     & \ul{0.797}     & \ul{0.823}     & \ul{0.806}     & \ul{0.823}     \\
                                      & Ours            & \textbf{0.825}$_{\small{\pm\text{0.033}}}$ & 0.908$_{\small{\pm\text{0.016}}}$ & 0.948$_{\small{\pm\text{0.010}}}$ 
                                                        & \textbf{0.822}$_{\small{\pm\text{0.011}}}$ & \textbf{0.822}$_{\small{\pm\text{0.010}}}$ & \textbf{0.828}$_{\small{\pm\text{0.010}}}$ 
                                                        & \textbf{0.884}$_{\small{\pm\text{0.021}}}$ & \textbf{0.886}$_{\small{\pm\text{0.029}}}$ & \textbf{0.909}$_{\small{\pm\text{0.030}}}$ \\
\bottomrule
\end{tabular}
}
\caption{10-shot cross-domain experiment results, the best and second best results are in bold and underlined. FE, GR, GC, LI and PH represent the source datasets FEVER, GettingReal, GossipCop, LIAR and PHEME.}
\label{tab:results}
\vspace{-10pt}
\end{table*}

\noindent
\textbf{Datasets and Baselines}: To examine \ours performance, we adopt multiple source and target datasets. We follow~\cite{yue2022contrastive} and adopt FEVER (FE)~\cite{thorne2018fever}, GettingReal (GR)~\cite{meganrisdal2016}, GossipCop (GC)~\cite{shu2020fakenewsnet}, LIAR (LI)~\cite{wang2017liar} and PHEME (PH)~\cite{buntain2017automatically} as the source datasets. For the target domain,  we adopt CoAID~\cite{cui2020coaid}, Constraint~\cite{patwa2021fighting} and ANTiVax~\cite{hayawi2022anti}. Our na{\"i}ve baseline leverages few-shot target examples to fine-tune the source pretrained models. We also adopt state-of-the-art baselines from domain adaptation and few-shot learning methods for domain adaptive few-shot misinformation detection: CANMD, ACLR, ProtoNet and MAML~\cite{finn2017model, snell2017prototypical, lin-etal-2022-detect, yue2022contrastive}. Additionally, we select two large language models LLaMA and Alpaca to evaluate few-shot in-context learning (ICL) and parameter-efficient fine-tuning (PEFT) performance in misinformation detection~\cite{touvron2023llama, alpaca}.

\noindent
\textbf{Implementation}: We follow the preprocessing pipeline as in~\cite{yue2022contrastive}. Specifically, we translate special symbols (e.g., emojis) back into English, tokenize hashtags, mentions and URLs and remove special characters from the input. We use the 10-shot setting (i.e., $k = 10$), the model is trained using AdamW optimizer with $0.01$ weight decay and no warm-up, where we sample $3$ source tasks and perform $3$ updates in inner-loop optimization. Then, we compute the meta loss and evaluate task similarity scores before rescaling the meta gradients with temperature $\tau$. All our main experiments are repeated 3 times, we select the best model with the validation set for final evaluation. Hyperparameter selection (e.g., inner and outer learning rates, temperature etc.) and implementation details are provided in \Cref{sec:implementation}.

\subsection{Main Results}
% supervised results
\noindent
\textbf{Supervised results}: We first report \emph{supervised} results on all datasets in \Cref{tab:supervised}. The upper and lower parts of the table report the source and target dataset performance respectively. We observe the following: (1)~overall, the performance on statements and posts achieves better performance than news articles. An example can be found on GossipCop (News) with 0.776 BA, compared to 0.863 on PHEME (Social network posts); (2)~for disproportionate label distributions (e.g., CoAID), the BA metric reduces drastically compared to other metrics, indicating the difficulty of training fair models on unfair distributions. For instance, BA is circa $10\%$ lower than accuracy and F1 on CoAID; (3)~the RoBERTa model only achieves an average BA of 0.778 on source datasets, which suggests that transferring knowledge from the source to the target datasets can be a challenging task.

% few-shot results and summary
\noindent
\textbf{Few-shot adaptation results}: The few-shot cross-domain experiments (10-shot) on all source-target combinations are presented in \Cref{tab:results}. In the table, rows represent the source datasets while the columns represent the target datasets. We include the adaptation methods in each row, while the metrics are reported in the columns under target datasets. For \ours, we report the mean results in the table and provide the standard deviation values using the $\pm$ sign. For convenience, the best results are marked in bold and the second best results are underlined. We observe: (1)~adapting to the COVID domain is a non-trivial task upon dissimilar source-target label distributions. In the example of GossipCop $\rightarrow$ CoAID, baseline methods show BA values to be slightly higher than 0.6 (despite high accuracy and F1), suggesting that the model predicts the majority class with a much higher likelihood; (2)~by learning domain-invariant feature representation, baseline methods like CANMD and ACLR can improve the adaptation results compared to the na{\"i}ve baseline. For instance, CANMD achieves over $42.3\%$ and $21.7\%$ average improvements on BA for Constraint and ANTiVax; (3)~using the meta adaptation approach, \ours significantly outperform all baselines in the BA metric. For instance, \ours outperforms the best-performing baseline MAML $7.4\%$ in BA on ANTiVax, with similar trends to be found in accuracy and F1. Specifically for CoAID (with over $90\%$ positive labels), improvements on BA demonstrates that \ours can learn fair features for improved detection results despite slightly worse accuracy and F1 results. In summary, the results in \Cref{tab:results} show that \ours is particularly effective in adapting early misinformation detection systems using limited target examples. In contrast to the baseline models, \ours can achieve significant improvements on all metrics across source-target dataset combinations. In the case of large domain discrepancy, \ours demonstrates superior performance by exploiting the few-shot target examples with second-order dynamics and similarity-based adaptive learning.

\begin{table}[t]
\small
\centering
\centerline{
\begin{tabular}{@{}llccc@{}}
\toprule
\textbf{Setting}                     & \textbf{Dataset} & \textbf{\; BA $\uparrow$ \;} & \textbf{\; Acc. $\uparrow$ \;} & \textbf{\; F1 $\uparrow$ \;} \\ \midrule
\multirow{3}{*}{\textbf{LLaMA-ICL}}  & CoAID            & 0.500                 & 0.906                   & 0.951              \\
                                     & Constraint       & 0.500                 & 0.523                   & 0.687              \\
                                     & ANTiVax          & 0.500                 & 0.664                   & 0.798              \\ \cmidrule(l){2-5}
\multirow{3}{*}{\textbf{Alpaca-ICL}} & CoAID            & 0.515                 & 0.908                   & 0.952              \\
                                     & Constraint       & 0.537                 & 0.559                   & 0.704              \\
                                     & ANTiVax          & 0.528                 & 0.681                   & 0.806              \\ \midrule
\multirow{3}{*}{\textbf{LLaMA-FT}}   & CoAID            & 0.749                 & 0.874                   & 0.928              \\
                                     & Constraint       & 0.724                 & 0.721                   & 0.718              \\
                                     & ANTiVax          & 0.742                 & 0.756                   & 0.811              \\ \cmidrule(l){2-5}
\multirow{3}{*}{\textbf{Alpaca-FT}}  & CoAID            & 0.766                 & 0.818                   & 0.892              \\
                                     & Constraint       & 0.688                 & 0.686                   & 0.689              \\
                                     & ANTiVax          & 0.767                 & 0.779                   & 0.828              \\ \midrule
\multirow{3}{*}{\textbf{\ours}}      & CoAID            & 0.825                 & 0.908                   & 0.948              \\
                                     & Constraint       & 0.822                 & 0.822                   & 0.828              \\
                                     & ANTiVax          & 0.884                 & 0.886                   & 0.909              \\ \bottomrule
\end{tabular}
}
\caption{Comparison to large language models.}
\label{tab:llm_comparison}
\vspace{-10pt}
\end{table}

% comparison to llm
\noindent
\textbf{Comparison to large language models}: To further demonstrate the effectiveness of \ours in domain adaptive misinformation detection, we compare our \ours with state-of-the-art large language models (LLMs). In particular, we adopt LLaMA-7B and Alpaca-7B and perform both few-shot in-context learning (i.e., ICL) and parameter-efficient fine-tuning (i.e., FT) on target datasets, with results presented in \Cref{tab:llm_comparison}. We notice: (1)~despite the significant increase in model parameters (from $\sim$0.1B to 7B), LLMs can still fail to distinguish misinformation without further tuning. For instance, LLaMA consistently scores 0.5 in BA by predicting the majority class in in-context learning. (2)~Fine-tuning LLMs can significantly improve the performance in misinformation detection. With PEFT tuning, Alpaca achieves $40.7\%$ performance improvement in BA across target datasets conditioned only on the few-shot target examples. (3)~With the proposed \ours, smaller language models like RoBERTa are capable of outperforming fine-tuned large language models. On average, \ours outperforms LLaMA-PEFT and Alpaca-PEFT by $14.3\%$ and $14.1\%$ in the BA metric on target datasets. The results suggest that \ours is both effective and efficient in early misinformation detection by combining out-of-domain knowledge and meta adaptation.

\begin{table}[t]
\small
\centering
\begin{tabular}{@{}llccc@{}}
\toprule
\textbf{Setting}                  & \textbf{Dataset} & \textbf{\; BA $\uparrow$ \;} & \textbf{\; Acc. $\uparrow$ \;} & \textbf{\; F1 $\uparrow$ \;} \\ \midrule
\multirow{3}{*}{\textbf{0-shot}}  & CoAID            & 0.551                 & 0.868                   & 0.925              \\
                                  & Constraint       & 0.567                 & 0.585                   & 0.705              \\
                                  & ANTiVax          & 0.528                 & 0.590                   & 0.689              \\ \cmidrule(l){2-5}
\multirow{3}{*}{\textbf{1-shot}}  & CoAID            & 0.594                 & 0.450                   & 0.588              \\
                                  & Constraint       & 0.664                 & 0.662                   & 0.656              \\
                                  & ANTiVax          & 0.627                 & 0.616                   & 0.645              \\ \cmidrule(l){2-5}
\multirow{3}{*}{\textbf{5-shot}}  & CoAID            & 0.728                 & 0.904                   & 0.949              \\
                                  & Constraint       & 0.799                 & 0.796                   & 0.792              \\
                                  & ANTiVax          & 0.700                 & 0.721                   & 0.776              \\ \cmidrule(l){2-5}
\multirow{3}{*}{\textbf{10-shot}} & CoAID            & 0.825                 & 0.908                   & 0.948              \\
                                  & Constraint       & 0.822                 & 0.822                   & 0.828              \\
                                  & ANTiVax          & 0.884                 & 0.886                   & 0.909              \\ \cmidrule(l){2-5}
\multirow{3}{*}{\textbf{15-shot}} & CoAID            & 0.876                 & 0.938                   & 0.964              \\
                                  & Constraint       & 0.844                 & 0.844                   & 0.870              \\
                                  & ANTiVax          & 0.865                 & 0.850                   & 0.872              \\ \bottomrule
\end{tabular}
\caption{Sensitivity on the number of target examples.}
\label{tab:fewshot_sensitivity}
\vspace{-10pt}
\end{table}

\begin{table*}[t]
\small
\centering
\begin{tabular}{@{}lccccccccc@{}}
\toprule
                & \multicolumn{3}{c}{\textbf{CoAID (2020)}}         & \multicolumn{3}{c}{\textbf{Constraint (2021)}}  & \multicolumn{3}{c}{\textbf{ANTiVax (2022)}}         \\ \cmidrule(l){2-10} 
\textbf{Metric} & \textbf{BA $\uparrow$} & \textbf{Acc. $\uparrow$} & \textbf{F1 $\uparrow$} & \textbf{BA $\uparrow$} & \textbf{Acc. $\uparrow$} & \textbf{F1 $\uparrow$} & \textbf{BA $\uparrow$} & \textbf{Acc. $\uparrow$} & \textbf{F1 $\uparrow$} \\
\midrule
\ours                     & 0.825 & 0.908 & 0.948 & 0.822 & 0.822 & 0.828 & 0.884 & 0.886 & 0.909 \\
\quad w/o Task Similarity & 0.811 & 0.909 & 0.948 & 0.807 & 0.805 & 0.806 & 0.862 & 0.853 & 0.880 \\
\quad w/o Adaptive LR     & 0.801 & 0.919 & 0.954 & 0.811 & 0.808 & 0.801 & 0.860 & 0.860 & 0.885 \\
\quad w/o 2nd-order Grads & 0.789 & 0.928 & 0.960 & 0.800 & 0.794 & 0.774 & 0.843 & 0.843 & 0.871 \\
\bottomrule
\end{tabular}
\caption{Ablation study results.}
\label{tab:ablation}
\vspace{-10pt}
\end{table*}

% robustness
\textbf{Robustness study}: We also evaluate the sensitivity of \ours with respect to the number of few-shot examples. In particular, we select the number from 0 to 15 and use \ours to perform the adaptation. The results are averaged across the source datasets and reported in \Cref{tab:fewshot_sensitivity}. Note that for 0-shot experiments, we train the models on source datasets and directly evaluate on target data. We observe the following: (1)~as expected, the adaptation performance improves rapidly with increasing number of few-shot examples. (2)~Surprisingly, we observe performance drops in accuracy and F1 on CoAID when increasing the few-shot number from 0 to 1. This suggests that the large domain discrepancy between both domains may reduce the effectiveness of \ours; (3)~Overall, the magnitude of improvements grows rapidly at first and then slowly plateaus as we increase the few-shot number. The largest improvements can be found from 1-shot to 5-shot setting, where the BA results improve by $18.1\%$ on average. In sum, we observe that even limited number of target examples can improve the target domain performance with \ours. We provide additional hyperparameters sensitivity analysis in \Cref{sec:additional_result}.

% ablation
\noindent
\textbf{Ablation study}: We evaluate the effectiveness of the proposed components in \ours. In particular, we remove the proposed similarity-based gradients rescaling (w/o Task Similarity), adaptive learning rate (w/o Adaptive LR) and second-order optimization method (w/o 2nd-order Grads) in order and observe the performance changes. Experiment results on target datasets are averaged across the source datasets and are reported in \Cref{tab:ablation}. For all components, we observe performance drops on BA when removed from \ours\footnote{In the case of CoAID, Acc and F1 are less reliable evaluation metrics due to the imbalanced distribution of labels.}. For example, the performance of \ours reduces by $2.0\%$ and $2.3\%$ in BA when we remove task similarity and adaptive LR consecutively. We further replace \ours with first-order approximation, which results in a performance drop of $3.9\%$ in the BA metric. Overall, the results suggest that the proposed components are effective for domain adaptive few-shot misinformation detection.
\section{Conclusion}
In this paper, we explore meta learning in domain adaptive few-shot misinformation detection. We propose \ours, a meta learning-based approach for cross-domain few-shot adaptation. \ours is the first to leverage few-shot target examples for exploiting the source domain knowledge under the guidance of limited target examples. Experiment results demonstrate the effectiveness of our method by achieving promising results on multiple challenging source-target dataset combinations, where \ours can significant outperform the state-of-the-art baselines and large language models by a significant margin.

\section{Limitations}
Despite introducing meta learning for domain adaptive few-shot misinformation detection, we have not discussed the setting of cross-domain adaptation with multiple source datasets to further improve the performance for identifying early-stage misinformation. Due to the lack of early-stage misinformation data, we limit our choice of the target domain to COVID-19, which may hinder the generalization of the proposed method to other domains. Additionally, the proposed method does not leverage efficient approximation or first-order meta learning methods to reduce the computational costs in training \ours. As such, we plan to explore multi-source few-shot misinformation detection via efficient meta learning as future work.

% \section*{Acknowledgments}

% This research is supported in part by the National Science Foundation under Grant No. IIS-2202481, CHE-2105005, IIS-2008228, CNS-1845639, CNS-1831669. The views and conclusions contained in this document are those of the authors and should not be interpreted as representing the official policies, either expressed or implied, of the U.S. Government. The U.S. Government is authorized to reproduce and distribute reprints for Government purposes notwithstanding any copyright notation here on.

% Entries for the entire Anthology, followed by custom entries
\bibliography{anthology,custom}
\bibliographystyle{acl_natbib}

\clearpage
\appendix
\section{Implementation}
\label{sec:implementation}

\subsection{Datasets}
We adopt FEVER~\cite{thorne2018fever}, GettingReal~\cite{meganrisdal2016}, GossipCop~\cite{shu2020fakenewsnet}, LIAR~\cite{wang2017liar} and PHEME~\cite{buntain2017automatically} as the source datasets. For the target domain,  we adopt three COVID datasets: ANTiVax (2022)~\cite{hayawi2022anti}, CoAID (2020)~\cite{cui2020coaid} and Constraint (2021)~\cite{patwa2021fighting}.

In the following, we present the details of our source and target datasets:
\begin{enumerate}[leftmargin=15pt, itemsep=3px, nolistsep]
  \item \textbf{FEVER} is a publicly available dataset for fact verification. FEVER consists of modified claims from Wikipedia without knowledge of the claims~\cite{thorne2018fever}.
  \item \textbf{GettingReal} is a fake news dataset from Kaggle (Getting Real about Fake News). GettingReal contains text and metadata scraped from online resources~\cite{meganrisdal2016}.
  \item \textbf{GossipCop} provides a fake news data from news content, social network, and dynamic iteratcions. GossipCop is adopted from the FakeNewsNet dataset~\cite{shu2020fakenewsnet}.
  \item \textbf{LIAR} is a publicly available dataset of fact verification. The provided statements are collected from politifact.com with analysis report and links to sources~\cite{wang2017liar}.
  \item \textbf{PHEME} contains tweets rumours and non-rumours in certain breaking events (e.g., Germanwings crash). PHEME provides online interactions and structure of the tweets~\cite{buntain2017automatically}.
  \item \textbf{CoAID} is a dataset with COVID-19 misinformation. CoAID provides fake news on websites and social platforms as well as user iteractions under such sources~\cite{cui2020coaid}.
  \item \textbf{Constraint} is a shared taks on COVID-19 fake news detection. It contains over 10k annotated social media posts and articles of real and fake news on COVID-19~\cite{patwa2021fighting}.
  \item \textbf{ANTiVax} is a novel dataset with over 15k COVID-19 vaccine-related tweets and annotations for vaccine misinformation detection~\cite{hayawi2022anti}.
\end{enumerate}

Details of the above datasets are provided in \Cref{tab:datasets}, where \textbf{Neg.} and \textbf{Pos.} are the proportion of misinformation and valid information in the dataset (i.e., label distribution). \textbf{Len} represents the average token length of text and \textbf{Content} denotes the source type of the text (e.g., statement, news or social posts). Notice that CoAID is largely imbalanced with over $90\%$ positive examples.

\begin{table}[h]
\small
\centering
\begin{tabular}{@{}lcccc@{}}
\toprule
\textbf{Datasets}    & \textbf{Neg.} & \textbf{Pos.} & \textbf{Len} & \textbf{Content} \\ \midrule
\textbf{FEVER}       & 29.6\%        & 70.4\%        & 9.4          & Statement        \\
\textbf{GettingReal} & 8.8\%         & 91.2\%        & 738.9        & News             \\
\textbf{GossipCop}   & 24.2\%        & 75.8\%        & 712.9        & News             \\
\textbf{LIAR}        & 44.2\%        & 55.8\%        & 20.2         & Statement        \\
\textbf{PHEME}       & 34.0\%        & 66.0\%        & 21.5         & Social Network   \\
\textbf{CoAID}       & 9.7\%         & 90.3\%        & 54.0         & News / Statement \\
\textbf{Constraint}  & 47.7\%        & 52.3\%        & 32.7         & Social Network   \\
\textbf{ANTiVax}     & 38.3          & 61.7\%        & 26.2         & Social Network   \\ \bottomrule
\end{tabular}
\caption{Details of the involved datasets.}
\label{tab:datasets}
\end{table}

\subsection{Baseline Methods}
As a na{\"i}ve baseline, we pretrained the misinformation detection model on the source dataset and fine-tune with the few-shot examples, followed by evaluation on the test set. We additionally adopt the following state-of-the-art baselines for domain adaptive few-shot misinformation detection:
\begin{enumerate}[leftmargin=15pt, itemsep=3px, nolistsep]
\item \textbf{Contrastive Adaptation Network for Misinformation Detection (CANMD)} proposes to use label correction in the pseudo-labeling process to generate labeled target examples. Then, contrastive learning is applied to learn domain-invariant and class-separating features. Therefore, we adapt CANMD by including the few-shot target examples in the training process as in our setting, we select the best results from the original CANMD and our adaptation~\cite{yue2022contrastive}.
\item \textbf{Adversarial Contrastive Learning for low-resource rumor detection (ACLR)} leverages language alignment and contrastive learning to improve corss-domain misinformation detection performance. ACLR also introduces adversarial augmentation to enhance the robustness of few-shot rumor detection. We replace the original graph convolution networks with our base transformer model for content-based misinformation detection~\cite{lin-etal-2022-detect}.
\item \textbf{Prototypical Networks for Few-shot Learning (ProtoNet)} designs prototypical networks for few-shot classification problems. ProtoNet projects sample features to a metric space and perform inference by computing distances to prototypes of each class. We adapt ProtoNet to meta adaptation framework by adopting the same label space and the base transformer model as encoder for domain-adaptive few-shot misinformation detection ~\cite{snell2017prototypical}.
\item \textbf{Model-Agnostic Meta-Learning (MAML)} is the first to leverage second-order derivatives for few-shot learning. MAML first update model parameters upon sampled tasks, followed by computing the meta loss and derive the second-order gradients w.r.t. the original parameters. Similarly, We adapt MAML to our meta adaptation framework with the homogeneous label space acorss tasks and the transformer encoder for misinformation detection~\cite{finn2017model}.
\end{enumerate}

\begin{figure}[t]
    \centering
    \includegraphics[trim=0 0 1cm 0, clip, width=0.85\linewidth]{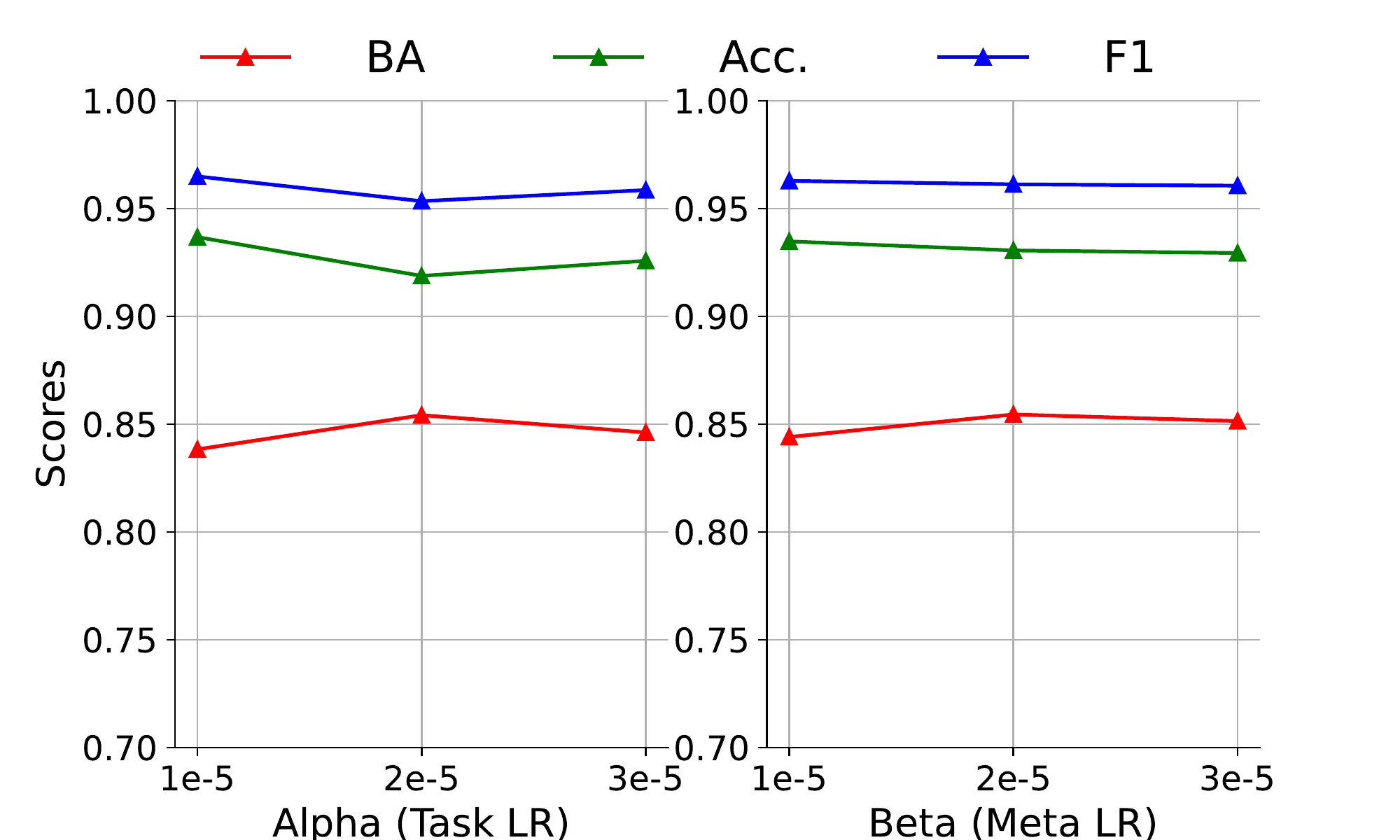}
    \caption{Sensitivity of the initial learning rates.}
    \label{fig:sensitivity_lr}
    \vspace{-10pt}
\end{figure}

\subsection{Implementation Details}
For our evaluation method, we follow the previous works~\cite{kou2022hc, li2021multi, yue2022contrastive} and split the datasets into training, validation, and test sets with the ratio of 7:2:1. If the dataset provides a default split, we directly use the provided split sets. The validation sets are used for label correction in~\cite{yue2022contrastive} and constructing few-shot examples and saving the best model in training. For our few-shot adaptation setting, we select the first $k$ examples (10-shot as default) from the original validation set and used the remaining examples for validation. We use accuracy and F1 score for evaluation. The balanced accuracy (BA) is additionally introduced to evaluate the adaptation performance in both classes equally, BA is defined as the mean of sensitivity and specificity.

For the na{\"i}ve baseline, we train the base model on the source dataset using AdamW optimizer (learning rate 1e-5) without warm-up. Then, the model is fine-tuned on the few-shot examples under the same training condition. For other baseline methods, we use the original implementation (if provided) and follow the original hyperparameter configuration. Otherwise we reimplement the baseline methods and adopt the identical training pipeline (AdamW with 1e-5 learning rate and no warm-up). Few-shot learning baselines (i.e., ProtoNet \& MAML) are adapted to the domain adaptive few-shot learning framework by using the few-shot target examples as query set. For the large language models, in-context learning is performed using the generative and perplexity-based approach based on~\cite{wu2023openicl}, while fine-tuning is performed using low-rank adaptation as in~\cite{hu2021lora}. In the \ours experiments, we adopt $3$ as the number of tasks $n$ and update the task specific model $\bm{\phi}$ for $3$ times in inner-level optimization, notice for each task we start from the original parameter set $\bm{\theta}$. To train \ours, we use a training batch size of $4$ in each task and the learning rates (both $\alpha$ and $\beta$) are selected from $[1e-5, 2e-5, 3e-5]$. The temperature hyperparameter is selected from $[1, 0.1, 0.01]$ (See sensitivity analysis in \Cref{sec:additional_result}). We perform training with $500$ or $1000$ meta iterations and validate the model every $50$ iterations, the best model is used for evaluation on the test sets.

\begin{figure}[t]
    \centering
    \includegraphics[trim=0 0 1cm 0, clip, width=0.85\linewidth]{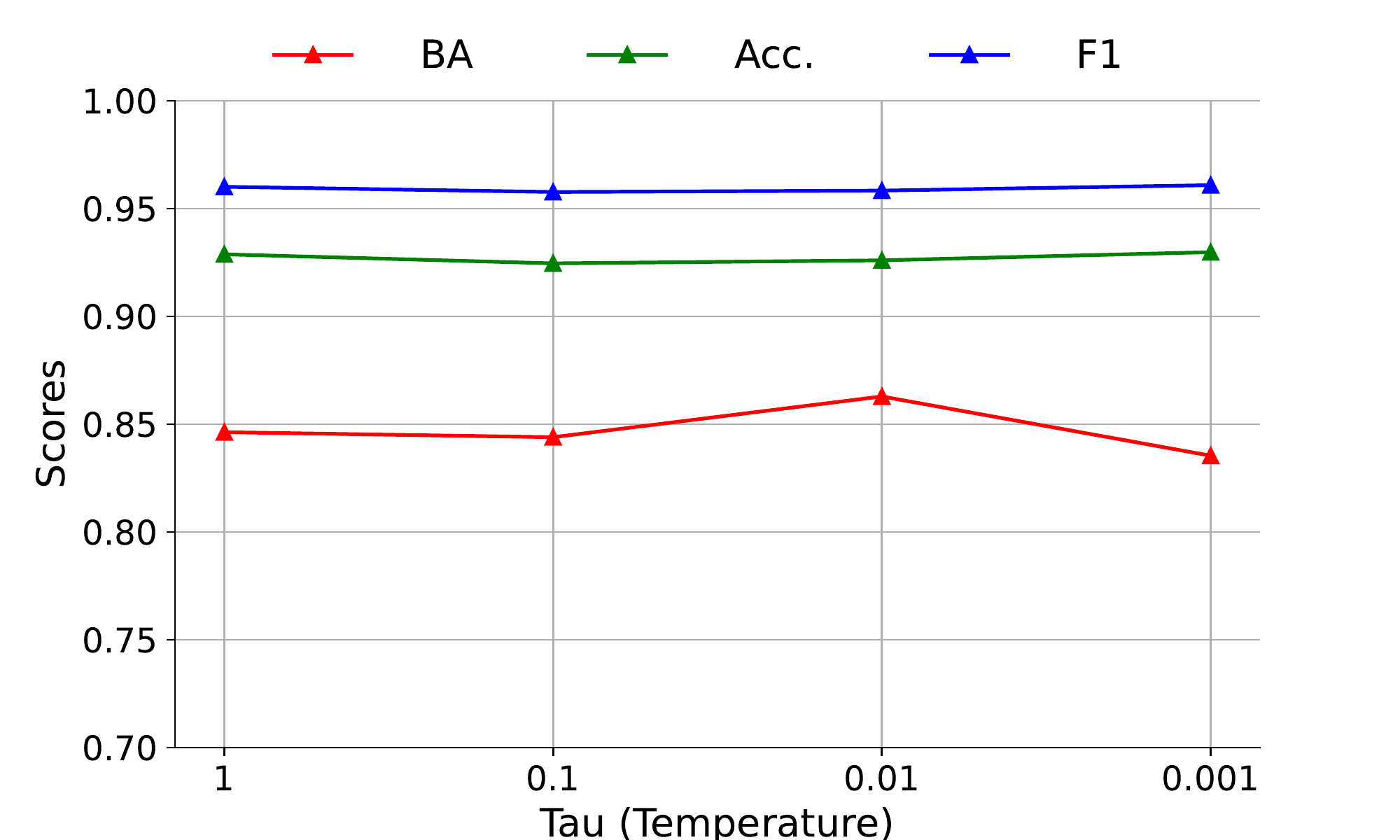}
    \caption{Sensitivity of temperature values.}
    \label{fig:sensitivity_temp}
    \vspace{-10pt}
\end{figure}

\section{Additional Results}
\label{sec:additional_result}

\noindent
\textbf{Sensitivity Analysis of Initial Learning Rates}: We study the sensitivity of $\alpha$ and $\beta$ on the CoAID dataset, the best results (averaged across source datasets) are presented in \Cref{fig:sensitivity_lr}. Overall, we observe that the performance of \ours is insensitive to the changes of both learning rates. Interestingly, we notice that accuracy values are negatively correlated with the BA scores, suggesting that accuracy may not be an ideal metric for data distributions with disproportionate label classes. 

\noindent
\textbf{Sensitivity Analysis of Temperature Values}: We also study the sensitivity of $\tau$ in gradient rescaling on CoAID, we present the best results (averaged across source datasets) in \Cref{fig:sensitivity_temp}. In short, the performance of \ours first increases and then starts to reduce, with the best results ocurring at $\tau = 0.01$, indicating the effectiveness of similarity-based gradients rescaling. Overall, the proposed \ours is robust to the hyperparameters and outperforms the baseline methods consistently.

\end{document}